\documentclass[runningheads]{llncs}
\usepackage[T1]{fontenc}
\usepackage{graphicx}
\usepackage{hyperref}       
\usepackage{url}            
\usepackage{booktabs}       
\usepackage{amsfonts}       
\usepackage{nicefrac}       
\usepackage{microtype}      
\usepackage{doi}
\usepackage{tabularx}
\usepackage{longtable}
\usepackage{xcolor}

\begin{document}
\title{CRRG-CLIP: Automatic Generation of Chest Radiology Reports and Classification of Chest Radiographs}
%
%
\author{Jianfei Xu\inst{1} \and Thanet Markchom\inst{2} \and
Huizhi Liang\inst{1}}

\authorrunning{J. Xu et al.}
%
\institute{Newcastle University, Newcastle upon Tyne NE4 5TG, United Kingdom \and
University of Reading, Reading RG6 6UR, United Kingdom
\email{mr.xujianfei@gmail.com}}
\maketitle              
\begin{abstract}
The complexity of stacked imaging and the massive number of radiographs make writing radiology reports complex and inefficient. Even highly experienced radiologists struggle to maintain accuracy and consistency in interpreting radiographs under prolonged high-intensity work. To address these issues, this work proposes the CRRG-CLIP Model (Chest Radiology Report Generation and Radiograph Classification Model), an end-to-end model for automated report generation and radiograph classification. The model consists of two modules: the radiology report generation module and the radiograph classification module. The generation module uses Faster R-CNN to identify anatomical regions in radiographs, a binary classifier to select key regions, and GPT-2 to generate semantically coherent reports. The classification module uses the unsupervised Contrastive Language–Image Pre-training (CLIP) model, addressing the challenges of high-cost labelled datasets and insufficient features. The results show that the generation module performs comparably to high-performance baseline models on BLEU, METEOR, and ROUGE-L metrics, and outperformed the GPT-4o model on BLEU-2, BLEU-3, BLEU-4, and ROUGE-L metrics. The classification module significantly surpasses the state-of-the-art model in AUC and Accuracy. This demonstrates that the proposed model achieves high accuracy, readability, and fluency in report generation, while multi-modal contrastive training with unlabelled radiograph-report pairs enhances classification performance.
The code can be accessed via the following link: \href{https://github.com/jianfeixu95/CSC8639}{https://github.com/jianfeixu95/CSC8639}

\end{abstract}

\section{Introduction}

Chest radiographs are commonly utilized in clinical disease screening and diagnosis because of their advantages of fast imaging and high definition \cite{Irvin2019}. However, due to the complex imaging characteristics and high-frequency use of chest radiographs, it is challenging for expert radiologists to accurately and consistently process and interpret the vast amount of complex information \cite{Robinson2016}. Studies show that 20\%-50\% of nodule diagnoses are missed or misdiagnosed on chest radiographs, and 3\%-6\% of cases result in serious clinical errors, even among the most experienced and distinguished radiologists \cite{Yan2019}. To solve these problems, researchers have explored the use of deep learning techniques for classifying chest radiographs \cite{Ma2020,Radford2021,Ziegelmayer2023}.

Recently, radiograph classification models have evolved from relying solely on radiographs to incorporating both radiographs and reports to enhance classification performance \cite{DallaSerra2022,Moon2022,Jacenkow2022}. Radiographs provide visual features of chest tissue structure, while radiology reports offer rich contextual information about diseases and patients. However, manually creating radiology reports demands significant human effort. Automatically generating radiology reports from radiographs presents a promising solution to this challenge.

Radiology report generation methods are predominantly driven by image processing techniques. They typically generate reports by extracting feature labels from radiographs and matching them to report templates. In contrast, methods based on natural language processing (NLP) leverage multi-task characteristics by using an image encoder to extract global features \cite{Wang2021,Wang2022} and a text generation model to decode and generate text \cite{Wang2021,Zhang2022,Yan2022}. While these NLP-based methods offer high content completeness with a focus on global features, they lack interpretability and ignore valuable local features \cite{Sloan2024}. This limitation can negatively impact the performance of downstream classification models.

Addressing these issues, this work proposes the Chest Radiology Report Generation and Radiograph Classification (CRRG-CLIP) Model, which is capable of automatically generating radiology reports and using radiographs and generated reports for classification. The proposed model consists of two modules: the radiology report generation module and the radiograph classification module. The radiology report generation module automatically extracts local visual features from radiographs, determines key regions, and generates a descriptive sentence for each key region to create a personalized report. This approach enables the generation process to focus on locally valuable regions, thereby improving the precision, fluency, and interpretability of the generated radiology reports. The radiograph classification module uses a self-supervised learning approach based on Contrastive Language–Image Pre-training (CLIP) \cite{Radford2021}. By leveraging self-supervised learning, the proposed classification module eliminates the need for costly labelled datasets, enhancing its accessibility, transferability and generalization. Overall, the CRRG-CLIP model not only provides effective diagnostic support for radiologists but also holds significant potential for disease diagnosis. The main research contributions are summarized as follows:

\begin{itemize}
   
    \item Propose a novel radiology report generation approach by integrating image processing techniques with natural language processing.

    \item Develop a cost-effective, self-supervised contrastive learning approach based on CLIP using an unlabeled chest radiograph-report dataset.

    \item Integrate radiology report generation capability and radiograph classification capability into an end-to-end model, enabling automated report generation and radiograph classification tasks.

    \item Conduct experiments to demonstrate that the generated reports closely resemble professional reports produced by radiologists and perform comparably to radiologists' reports in chest radiograph classification tasks.
    
\end{itemize}

\section{Related Work}

\paragraph{Object Detection} 

Ren et al. \cite{Ren2017} introduced the Faster R-CNN framework to accelerate object detection, which integrates the Region Proposal Network (RPN) with Fast R-CNN \cite{Girshick2015}. Kisilev et al. \cite{Kisilev2016} applied the Faster R-CNN framework to lesion region recognition in radiographs and proposed a method for semantic description of lesion regions. Ma et al. \cite{Ma2020} used enhanced Faster R-CNN to identify spinal cord lesion regions. The VGG used for feature extraction was replaced with ResNet-50, which enhanced the traditional Faster R-CNN model. Inspired by Ma et al. \cite{Ma2020}, this work will use the enhanced Faster R-CNN framework as the object detector. 

\paragraph{Report Generation}

The research on radiology report generation focuses on the encoder-decoder framework, showing the trend from the CNN-RNN neural network model to the Transformer \cite{Sloan2024}. Ni et al. \cite{Ni2020} solved the issue of poor fluency caused by the repetition of words and sentences, and a cross-modal retrieval method was introduced into the hierarchical CNN-RNN model. Xiong et al. \cite{Xiong2019} introduced a hierarchical Transformer model designed for generating radiology reports. Ziegelmayer et al. \cite{Ziegelmayer2023} assessed multi-modal GPT-4 to generate radiological reports. Despite the advancement, current report generation approaches often struggle with the inability to understand the context in radiographs, leading to low personalization in reports \cite{Kurisinkel2021}. The transparency of the generation process is low, and the poor interpretability of conclusions creates a trust gap between radiologists and the model \cite{Wu2023b}. Inspired by \cite{Alfarghaly2021,Ziegelmayer2023}, this work uses GPT-2 to generate radiology reports.

\paragraph{Radiograph Classification}

Supervised radiograph classification models heavily rely on high-cost datasets with accurate labels \cite{Xu2021}, resulting in failure to obtain high benefits. The unsupervised radiograph classification model can learn rich features from enormous low-cost unlabelled radiographs, which is widely sought after by scholars \cite{Wu2024}. However, the unsupervised model has poor transferability, and the model cannot recognize new categories. Mikolov et al. \cite{Mikolov2013} proposed a self-supervised model, and Zhang et al. \cite{Zhang2024} applied it in the field of radiographs and achieved good results. Chen et al. \cite{Chen2019} found that a self-supervised model utilizing contrastive learning can improve the classification performance of the model on chest radiographs. 

Self-supervised learning has also been leveraged for multi-modal learning. In radiography classification, images provide detailed visual features of the structure and shape of tissues and lesions, while reports contain helpful conclusions, symptoms and contexts. Radford et al. \cite{Radford2021} introduced Contrastive Language-Image Pre-Training (CLIP), which demonstrated outstanding performance across over 30 distinct visual classification tasks. Despite advances in radiograph classification, challenges remain in accurately classifying low-quality radiographs \cite{Ovacll2020} and effectively handling new or rare diseases with high performance \cite{Ito2021}.

\section{Chest Radiology Report Generation and Radiograph Classification Model}
\label{sec: methodology_section}

\begin{figure}[t]
    \centering
    \includegraphics[width=0.8\linewidth]{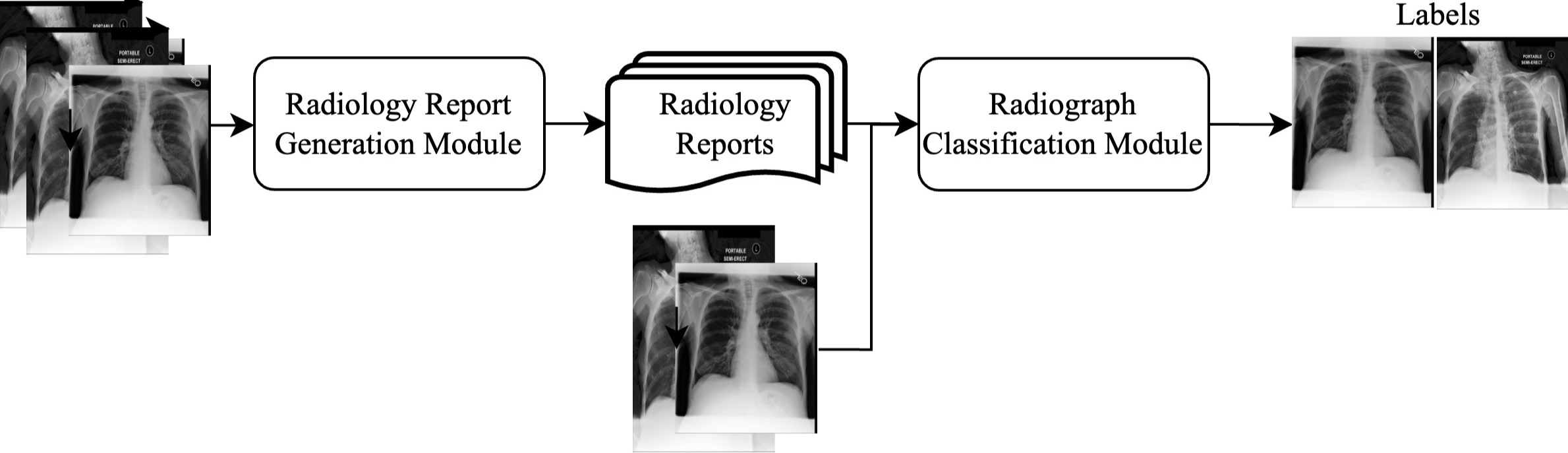}
    \caption{CRRG-CLIP Model Architecture.}
    \label{fig: high-level-model-architecture}
\end{figure}

\begin{figure}[t]
    \centering
    \includegraphics[width=0.8\linewidth]{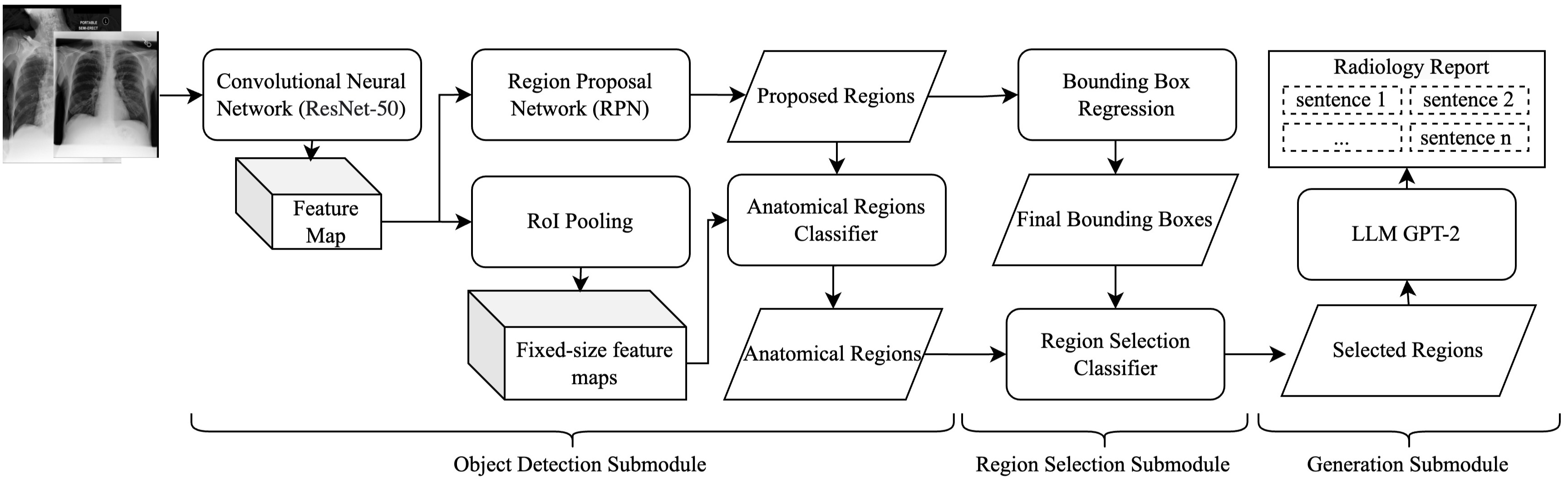}
    \caption{Radiology Report Generation Module Architecture.}
    \label{fig: report-generation-model-architecture}
\end{figure}

\begin{figure}[t]
    \centering
    \includegraphics[width=0.8\linewidth]{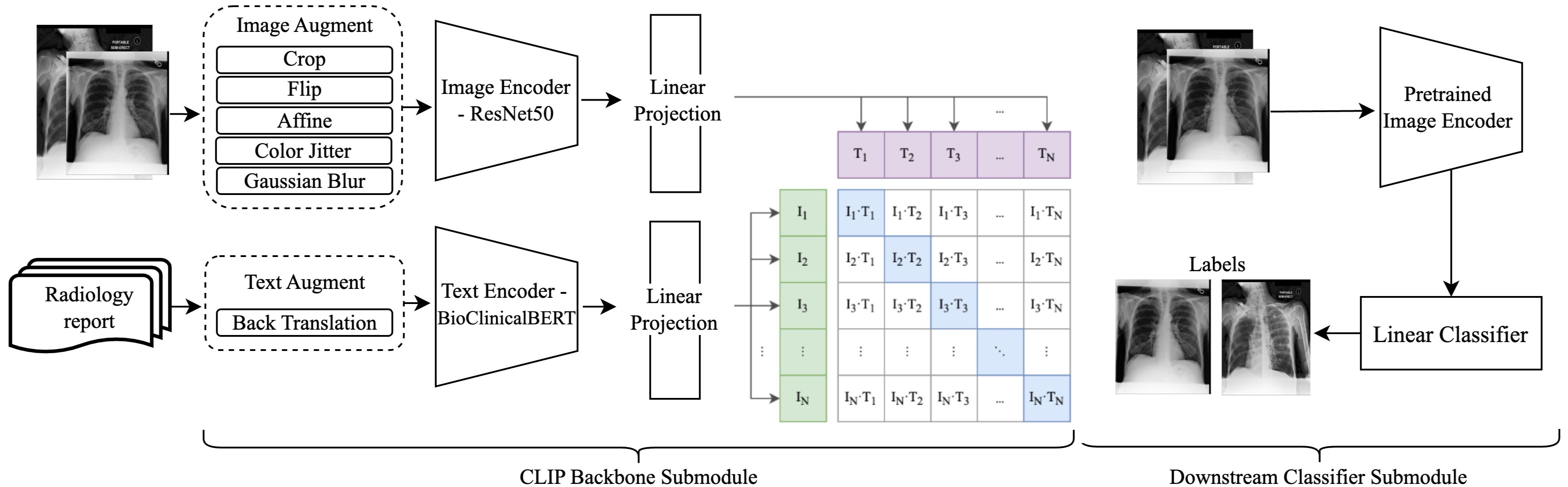}
    \caption{Radiograph Classification Module Architecture.}
    \label{fig: classification-model-architecture}
\end{figure}

The proposed Chest Radiology Report Generation and Radiograph Classification Model (CRRG-CLIP) (Figure \ref{fig: high-level-model-architecture}) consists of two parts: the radiology report generation (RRG) module and the radiograph classification (R-CLIP) module. RRG module (Figure \ref{fig: report-generation-model-architecture}) consists of an object detection submodule, a region selection submodule, and a generation submodule. The R-CLIP module (Figure \ref{fig: classification-model-architecture}) consists of a CLIP backbone, which includes both an image encoder and a text encoder, along with a downstream linear classifier submodule.

\subsection{Radiology Report Generation (RRG) Module}

\paragraph{Object Detection Submodule}

The Faster R-CNN model \cite{Ren2017} is adopted for object detection since it can accurately detect 29 anatomical regions \cite{Tanida2023} and get the boundary coordinates for each region \cite{Nasrullah2019}. When a radiograph is fed into the module, the image features are initially extracted using a ResNet-50 \cite{He2016} that has been pre-trained on ImageNet \cite{Deng2009}, and a feature map is generated. Then, on the one hand, a window is sliding over the feature map via the RPN \cite{Ren2017}, and RPN generates border coordinate predictions and target scores for multiple candidate regions. On the other hand, the ROI pooling layer adjusts the feature map to a fixed size, which avoids the complex operation of the subsequent classifier to deal with variable size and improves the processing efficiency of the network. Next, the candidate regions are classified using a classifier to obtain the anatomical region labels and bounding box coordinates. Finally, Bounding Box Regression \cite{Girshick2015} is used to obtain the offset of the border, and the border coordinates of the candidate regions are adjusted to improve the accuracy of the border around the target, and the final coordinates of the border are obtained. When classifying 29 anatomical regions, the output of the multi-class classifier is the probability distribution of the class of each anatomical region. For a single candidate region, the class corresponding to the highest probability score is the class of the candidate region. For multiple candidate regions of the same anatomical region, the candidate region corresponding to the highest probability score is the selected region of this class.

\paragraph{Region Selection Submodule}

When interpreting radiographs, radiologists select valuable areas based on their professional knowledge and experience to give diagnostic sentences \cite{Yan2022}. Inspired by this, a supervised binary classifier is used for region selection. The region selection classifier determines whether each valuable region is necessary to generate sentences by learning the attribute of whether the anatomical regions have annotated sentences. The classifier utilizes a fully connected feed-forward neural network with three layers, featuring input dimensions of 1024, 512, and 128. ReLU activation functions are applied between each fully connected layer. To evaluate the difference between predicted results and actual labels, a binary cross-entropy with logits loss function is employed.

\paragraph{Generation Submodule}

This work uses the fine-tuned GPT-2 \cite{Alec2019} (healx/gpt-2-pubmed-medium\footnote{\url{https://huggingface.co/healx/gpt-2-pubmed-medium}}) model to generate diagnostic sentences based on region features.

\subsection{Radiograph Classification (R-CLIP) Module}

\paragraph{CLIP Backbone Submodule}

In this submodule, the image encoder uses RestNet-50 to extract image features. The text encoder uses the fine-tuning BioClinicalBERT \cite{Yan2022} model (\verb|emilyalsentzer/Bio_ClinicalBERT|\footnote{\url{https://huggingface.co/emilyalsentzer/Bio\_ClinicalBERT}}) to extract text features. Image embeddings and text embeddings are each projected to the same size by a linear projection and aligned to the same embedding space. Then, the loss value, which includes Multi-View Supervised Contrastive Loss (MVS), Instance Contrastive Loss (ICL), and Triplet Contrastive Loss (TCL) \cite{You2023}, is utilized to guide the training process. This approach brings similar image-text pairs closer together in the feature space while pushing dissimilar pairs further apart \cite{Radford2021}. 

\paragraph{Downstream Classifier Submodule}

The downstream classifier is a simple PyTorch-based module consisting of a single fully connected linear layer with an input size of 224 and an output size of 1. It performs a linear transformation on the input features, generating logits for the target class. A sigmoid activation function is applied externally for binary classification.

\subsection{Training Procedure}

The training process for the model is split into two phases. The initial phase focuses on training the radiology report generation module, while the subsequent phase involves training the radiograph classification module. During the training of the radiology report generation module, the object detection submodule is first trained, so that the model can identify 29 key anatomical regions in the radiograph. The region selection submodule is then trained so that the model can determine the most valuable bounding box for generating the report. Finally, the generation submodule is trained to generate reports according to the image features in the bounding box. During the training of the radiograph classification module, firstly, the CLIP backbone submodule is trained, so that the image and text encoders can extract features respectively, and understand the relationship between radiographs and reports. The final step involved training a downstream classifier submodule that can classify radiographs.

\section{Experiments}


\subsection{Datasets}

To conduct experiments, this work uses four datasets:

\begin{itemize}
    \item \textbf{MIMIC-CXR Database} \cite{Johnson2019a} contains data from 227,835 chest radiology reports in TXT format with radiographs in DICOM format. Each report includes various sections. This study only focuses on the 'FINDINGS' section, representing the radiologists' diagnostic results. This helps avoid the impact of low-quality and incoherent erased patient information in other sections due to privacy protection \cite{Visser2020}. In our work, the reports were extracted and used to train the multimodal classification model, while the DICOM-format radiographs, which are large and challenging to process, were not utilized.

    \item \textbf{MIMIC-CXR-JPG Database} \cite{Johnson2019b} is derived from the database of the MIMIC-CXR. The DICOM format image file was converted to JPG format and the unstructured report was converted to structured disease labels. In our work, JPG format radiographs were used as the image data source.

    \item \textbf{Chest ImaGenome Dataset} \cite{Wu2021} is also derived from the MIMIC-CXR dataset and annotated in more detail. Each radiograph contains bounding boxes labelled as normal or abnormal tissues. Radiologists described the prominent features of each bounding box with sentences, indicating possible disease names. The final diagnosis report for each radiograph was generated from these sentences. In our work, local region coordinates in the dataset were employed to train the object detection submodule, the correspondence between local regions and diagnostic sentences was used to train the region selection submodule, and high-value local regions were employed as input for the report generation submodule to produce diagnostic sentences.

    \item \textbf{RSNA Pneumonia Dataset} \cite{RSNA2018} contains chest radiographs in DICOM format, labels for pneumonia, and other metadata. In the work, this dataset was used to train and evaluate the downstream linear classification submodule.
    
\end{itemize}

To form a complete dataset for training and evaluating the proposed model, the reports from the MIMIC-CXR Database, the radiographs from the MIMIC-CXR-JPG Database, and the scene graph JSON files from the Chest ImaGenome Dataset were matched using the ID fields \verb|subject_id|, \verb|study_id|, and \verb|image_id| for each radiology report.

\subsection{Preprocessing}

The radiology report generation module uses the dataset partitioning rules of the Chest ImaGenome Dataset. The CLIP backbone submodule in the radiograph classification module also uses the same partitioning. The downstream classifier submodule of the radiograph classification module uses data from the RSNA Pneumonia Dataset, partitioned into training, validation, and test sets in a 7:1.5:1.5 ratio. Due to GPU limitations, experiments were conducted on a reduced dataset of 10,000 sampled images (3.70\% of the dataset) and their associated reports, maintaining the same partition ratio.

The images from the MIMIC-CXR dataset were resized to 512 pixels on the long side, with black padding added to the short side to reach 512 pixels. Random colour dithering was applied to the H channel, Gaussian noise was added, random translation and rotation were performed, and a normalization operation was carried out. For the RSNA Pneumonia dataset, each image was randomly cropped from the centre to 224 pixels, and the brightness (±10\%), contrast (±20\%), saturation (±20\%) and hue (±10\%) were randomly changed.

The 'FINDINGS' section of the report was extracted with newline symbols removed. Back Translation \cite{Visser2020} was applied for data augmentation using Helsinki-NLP's Marian Machine Translation model, translating English to Italian and back to generate semantically similar but differently expressed texts.

\subsection{Baselines}

Evaluation was performed on the RRG model for a radiology report generation task and the R-CLIP model for a radiograph classification task. Each model was compared with various baseline models, including high-performance and commercial models, as well as its variations. The details of the compared models for each task are as follows:

\begin{itemize}

    \item Radiology Report Generation 
    
    \begin{itemize}

        \item  \textbf{S\&T} \cite{Vinyals2015}: a high-performance model that uses CNN and LSTM to construct a neural and probabilistic framework for caption generation. 
        
        \item \textbf{ADAATT} \cite{Lu2017}: a high-performance model extensively used in the literature. This model utilizes adaptive attention, allowing the model to decide where to focus on image features during training. 
        
        \item \textbf{GPT-4o} \cite{gpt4o}: the advanced GPT-4o from OpenAI. This model is considered as a commercial model used in numerous real-world applications. 
        
        \item \textbf{RRG-base}: the proposed RRG model using only 1.35\% of the complete dataset, serving as a baseline. In this model, the imbalance between regions with and without text bounding boxes was ignored, as well as the text generation length limit during report generation.
        
        \item \textbf{RRG-opt}: the optimized RRG model using 3.70\% of the complete dataset. The average number of tokens per report was used as the maximum text generation length. Weights were assigned to regions with and without text bounding boxes, ensuring the model fairly considers both types of regions. The hyperparameter settings for both \textbf{RRG-base} and \textbf{RRG-opt} can be found in Appendix \ref{appendix: parameter_settings} (Tables \ref{tab: report_generation_model_hyperparameter_settings} and  \ref{tab: report_generation_model_training_parameters}).
    
    \end{itemize}

    \item Radiograph Classification
    
    \begin{itemize}
    
        \item \textbf{ConVIRT} \cite{Zhang2022}: a state-of-the-art model using ..... . For this model's image encoder, both default random initialization and weights pre-trained on ImageNet \cite{Deng2009} were used as the initial weights.
        
        \item \textbf{R-CLIP-base}: a baseline version of the proposed R-CLIP model. This model was trained using image-text pairs consisting of the Radiologist’s reports and chest radiographs from the dataset. 
        
        \item \textbf{R-CLIP-opt}: an optimized version of the proposed R-CLIP model. It was trained on image-text pairs, which included radiographs from the dataset and corresponding reports produced by the RRG module. The detailed hyperparameters and training parameters are provided in Appendix \ref{appendix: parameter_settings} (Tables \ref{tab: radiograph_classification_hyperparameters} and \ref{tab: radiograph_classification_training_params}).
    \end{itemize}

\end{itemize}

All models were trained on Google Colab T4 GPU.

\subsection{Evaluation Metrics}

The generated reports were evaluated by \textbf{BLEU-1} (for the consistency and accuracy), \textbf{BLEU-2, BLEU-3, and BLEU-4} (for the readability and fluency \cite{Papineni2002}), \textbf{METEOR} (for consistent meaning with the ground truth), \textbf{ROUGE-L} (for the semantic consistency \cite{Banerjee2005}), \textbf{CIDEr} (for overlap degree of generated text and reference text), and \textbf{TF-IDF} \cite{H.Gomaa2013} (for the similarity between texts \cite{Vedantam2015}). Meanwhile, \textbf{Accuracy} and \textbf{AUC} were used to evaluate the classification model.




\section{Results}

\subsection{Report Generation Results}

\begin{figure}[t]
    \centering
    \includegraphics[width=0.8\linewidth]{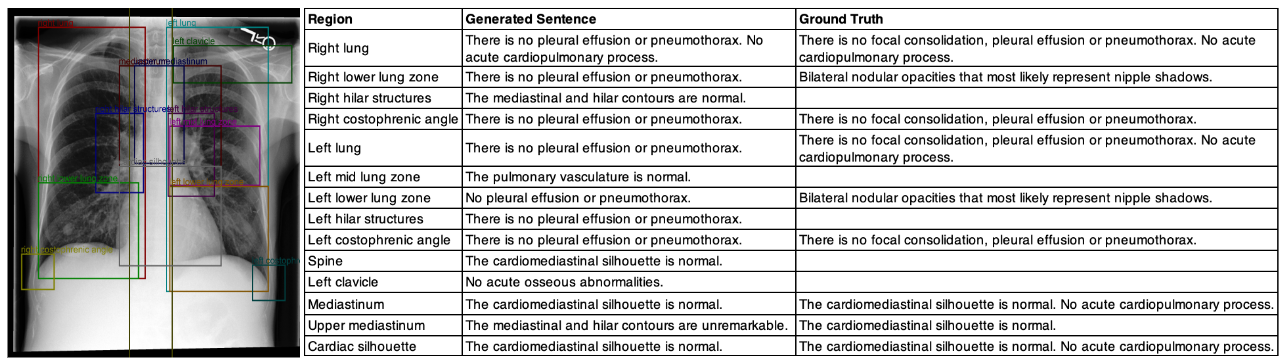}
    \caption{Example of a Result Generated by the Radiology Report Generation Module.}
    \label{fig: example_of_diagnostic_report_result}
\end{figure}

\begin{table}[t]
    \centering
    \scalebox{.85}{
    \begin{tabular}{lllllllll}
    \toprule
    \textbf{Model} & \textbf{Dataset} & \textbf{BLEU1} & \textbf{BLEU2} & \textbf{BLEU3} & \textbf{BLEU4} & \textbf{METEOR} & \textbf{ROUGEL} & \textbf{CIDEr} \\
    \midrule
    S\&T \cite{Vinyals2015} & 100\% & 0.299 & 0.184 & 0.121 & 0.084 & 0.124 & 0.263 & - \\
    ADAATT \cite{Lu2017} & 100\% & 0.299 & 0.185 & 0.124 & 0.088 & 0.118 & 0.266 & - \\
    GPT-4o & 3.70\% & 0.273 & 0.128 & 0.061 & 0.032 & 0.25 & 0.177 & - \\
    RRG-base (ours) & 1.35\% & 0.224 & 0.146 & 0.101 & 0.072 & 0.104 & 0.237 & 0.429 \\
    RRG-opt (ours) & 3.70\% & 0.241 & 0.157 & 0.108 & 0.078 & 0.109 & 0.239 & 0.513 \\
    \bottomrule
    \end{tabular}}
    \caption{Comparison of Radiology Report Generation Module. Note: Previous high-performance models did not compute CIDEr scores.}
    \label{tab: report_generation_model_results}
\end{table}

The proposed report generation approach shows significantly improved performance after optimization compared to the baseline model. The optimized report generation module performs similarly to the S\&T and ADAATT models and surpasses the GPT-4o in terms of fluency and readability, with the autogenerated reports closely resembling those of radiologists. Please refer to Figure \ref{fig: example_of_diagnostic_report_result} for an example of the result and Table \ref{tab: report_generation_model_results} for detailed experimental results.

\paragraph{Comparison with RRG-base} 

The baseline model RRG-base was trained using a 1.35\% dataset. By fine-tuning the baseline model and increasing the training data to 3.7\% of the dataset, the performance of the model was improved on all seven metrics. From the experimental results in Table \ref{tab: report_generation_model_results}, the BLEU-1 score increased by 7.59\%, BLEU-2 score by 7.53\%, BLEU-3 score by 6.93\%, BLEU-4 score by 8.33\%, METEOR score by 4.81\%, ROUGE-L score by 0.84\%, CIDEr score by 19.58\%, with an average improvement of 7.95\%.

\paragraph{Comparison with High-Performance Models}

Limited by GPU hardware equipment, the amount of data used in this experiment is small, only 3.7\% of the complete data set, but the performance of the trained model is close to that of previous high-performance models, S\&T and ADAATT, trained with the full dataset (Table \ref{tab: report_generation_model_results}), which proves that the model has strong performance under small samples, and the model can achieve better results if the experimental conditions are sufficient.

From Table \ref{tab: report_generation_model_results}, compared to the S\&T model, the BLEU-1 score of the model in this work is 24.07\% lower, BLEU-2 score is 17.20\% lower, BLEU-3 score is 12.04\% lower, BLEU-4 score is 7.69\% lower, METEOR score is 13.76\% lower, and ROUGE-L score is 10.04\% lower, with an average reduction of 14.13\%. In comparison with the ADAATT model, the BLEU-1 score of the model in this work is 24.07\% lower, BLEU-2 score is 17.83\% lower, BLEU-3 score is 14.82\% lower, BLEU-4 score is 18.82\% lower, METEOR score is 8.26\% lower, and ROUGE-L score is 11.30\% lower, with an average reduction of 14.85\%. In the comparison with the S\&T model and the ADAATT model, the reduction rates for BLEU-1 and BLEU-2 scores are higher than their respective average reduction rates (14.13\% and 14.85\%), while the reduction rates for BLEU-3, BLEU-4, METEOR, and ROUGE-L scores are lower than their respective average reduction rates.

\paragraph{Comparison with the Commercial Model}

Compared to the open-source GPT-2, GPT-4o features an increased layer count (48 to 120), 1 trillion parameters (up from 1.5 billion), and an improved attention mechanism, enhancing its performance on long texts and complex semantics. With the massive data of commercial companies and the parallel training of GPU with large computing power, the model effect has been further improved. However, only BLEU-1 and METEOR are slightly lower than GPT-4o in two of the seven evaluation dimensions, indicating that the generated report is better than GPT-4o in terms of readable ability and fluency, and slightly inferior to GPT-4o by 11.72\% in terms of vocabulary accuracy, and there is a gap in semantic effects such as text synonymous substitution and word derivation. Given that this model is trained on a limited dataset and constrained by the absence of open-source large language model architectures, this performance is also excellent.

\subsection{Radiograph Classification Results}

\begin{table}[t]
    \centering
    \begin{tabularx}{\textwidth}{lXlcc}
    \toprule
    \textbf{Model} & \textbf{Method} & \textbf{Dataset} & \textbf{AUC} & \textbf{Accuracy} \\
    \midrule
    ConVIRT \cite{Zhang2022} & Random Init & 1\% & 0.719 & - \\
            & ImageNet Init & 1\% & 0.831 & - \\
    R-CLIP-base (ours)    & Radiologist's report & 1\% & 0.852 & 0.788 \\
    R-CLIP-opt (ours)    & Generated report & 1\% & 0.848 & 0.780 \\
    \bottomrule
    \end{tabularx}
    \caption{Comparison of Radiograph Classification Module, Note: The ConVIRT model did not compute accuracy on the 1\% MIMIC-CXR database.}
    \label{tab: radiograph classification model results}
\end{table}

From Table \ref{tab: report_generation_model_results}, the performance of R-CLIP-base using radiologists' reports and R-CLIP-opt using reports generated by the generative module is similar. It shows that the proposed report generation approach produces reports comparable to those written by humans.  Additionally, the proposed classification approach outperformed ConVIRT. This indicates its superior performance compared to a state-of-the-art high-performance model.

\section{Conclusions}

In this work, the CRRG-CLIP model, composed of a radiology report generation module based on local features and a radiograph classification module based on multimodal features, is proposed. The radiology report generation module simulates the process of radiologists' interpretation of radiographs, which implements automatic, efficient and accurate segmentation of anatomical regions, extraction of valuable local tissue features, and generation of smooth and professional reports. It enhances the detailed description of the report by focusing on local anatomical features and improves the accuracy of diagnosis. It also solves the problems of low interpretability, poor readability and low fluency of previous models, and provides support for model tuning and abductive analysis of report content. The radiograph classification module enhances the type and quantity of features obtained, improving downstream classification performance by learning the consistency and differentiation of image-text features. The experimental results show that the reports generated by the radiology report generation module and the reports written by radiologists can achieve similar results in classification tasks, which proves that the radiology report generation module can assist radiologists in report writing. In addition, experiments also show that the proposed model can achieve strong performance on small datasets, which is close to the performance of previous high-performance models and exceeds the commercial model. Future work will focus on optimizing model performance through advanced architectures, experimenting with the full dataset, and incorporating human evaluation of generated reports.

\bibliographystyle{splncs04}
\bibliography{references}

\begin{thebibliography}{10}
\providecommand{\url}[1]{\texttt{#1}}
\providecommand{\urlprefix}{URL }
\providecommand{\doi}[1]{https://doi.org/#1}

\bibitem{Alec2019}
Alec, R., Jeffrey, W., Rewon, C., David, L., Dario, A., Ilya, S.: {Language Models are Unsupervised Multitask Learners | Enhanced Reader}. OpenAI Blog  \textbf{1}(8), ~9 (2019)

\bibitem{Alfarghaly2021}
Alfarghaly, O., Khaled, R., Elkorany, A., Helal, M., Fahmy, A.: {Automated radiology report generation using conditioned transformers}. Informatics in Medicine Unlocked  \textbf{24},  100557 (jan 2021)

\bibitem{Banerjee2005}
Banerjee, S., Lavie, A.: Meteor: An automatic metric for mt evaluation with improved correlation with human judgments. In: Proceedings of the acl workshop on intrinsic and extrinsic evaluation measures for machine translation and/or summarization. pp. 65--72 (2005)

\bibitem{Chen2019}
Chen, L., Bentley, P., Mori, K., Misawa, K., Fujiwara, M., Rueckert, D.: {Self-supervised learning for medical image analysis using image context restoration}. Medical Image Analysis  \textbf{58},  101539 (2019)

\bibitem{DallaSerra2022}
{Dalla Serra}, F., Jacenk{\'{o}}w, G., Deligianni, F., Dalton, J., O'Neil, A.Q.: {Improving Image Representations via MoCo Pre-training for Multimodal CXR Classification}. In: Lecture Notes in Computer Science (including subseries Lecture Notes in Artificial Intelligence and Lecture Notes in Bioinformatics). vol. 13413 LNCS, pp. 623--635. Springer Science and Business Media Deutschland GmbH (2022)

\bibitem{Deng2009}
Deng, J., Dong, W., Socher, R., Li, L.J., Li, K., Fei-Fei, L.: {ImageNet: A Large-Scale Hierarchical Image Database}. In: 2009 IEEE Conference on Computer Vision and Pattern Recognition, CVPR 2009. pp. 248--255 (2009)

\bibitem{Girshick2015}
Girshick, R.: {Fast R-CNN}. In: Proceedings of the IEEE International Conference on Computer Vision. pp. 1440--1448 (2015)

\bibitem{He2016}
He, K., Zhang, X., Ren, S., Sun, J.: {Deep residual learning for image recognition}. In: Proceedings of the IEEE Computer Society Conference on Computer Vision and Pattern Recognition. vol. 2016-Decem, pp. 770--778 (2016)

\bibitem{H.Gomaa2013}
H.Gomaa, W., {A. Fahmy}, A.: {A Survey of Text Similarity Approaches}. International Journal of Computer Applications  \textbf{68}(13),  13--18 (2013)

\bibitem{Irvin2019}
Irvin, J., Rajpurkar, P., Ko, M., Yu, Y., Ciurea-Ilcus, S., Chute, C., Marklund, H., Haghgoo, B., Ball, R., Shpanskaya, K., Seekins, J., Mong, D., Halabi, S., Sandberg, J., Jones, R., Larson, D., Langlotz, C., Patel, B., Lungren, M., Ng, A.: Chexpert: A large chest radiograph dataset with uncertainty labels and expert comparison. Proceedings of the AAAI Conference on Artificial Intelligence pp. 590--597 (2019)

\bibitem{Jacenkow2022}
Jacenkow, G., O'Neil, A.Q., Tsaftaris, S.A.: {Indication as Prior Knowledge for Multimodal Disease Classification in Chest Radiographs with Transformers}. In: Proceedings - International Symposium on Biomedical Imaging. vol. 2022-March. IEEE Computer Society (feb 2022), \url{https://arxiv.org/abs/2202.06076v1}

\bibitem{Johnson2019b}
Johnson, A., Lungren, M., Peng, Y., Lu, Z., Mark, R., Berkowitz, S., Horng, S.: {MIMIC-CXR-JPG - chest radiographs with structured labels (version 2.0.0)}. PhysioNet  (2019)

\bibitem{Johnson2019a}
Johnson, A.E., Pollard, T.J., Berkowitz, S.J., Greenbaum, N.R., Lungren, M.P., ying Deng, C., Mark, R.G., Horng, S.: {MIMIC-CXR, a de-identified publicly available database of chest radiographs with free-text reports}. Scientific Data  \textbf{6}(1) (2019)

\bibitem{RSNA2018}
Kaggle: Rsna pneumonia detection challenge (2018), \url{https://www.kaggle.com/competitions/rsna-pneumonia-detection-challenge}

\bibitem{Ito2021}
{Kentaro Ito, Keisuke Ogaki, Dongyi Xue, Jumpei Ukita, Seiwa Honda}, O.H.: {P16‐27: A deep learning approach using chest X‐ray data for screening drug‐induced interstitial lung disease}. Respirology  \textbf{26}(S3),  458--459 (2021)

\bibitem{Kisilev2016}
Kisilev, P., Sason, E., Barkan, E., Hashoul, S.: {Medical image description using multi-task-loss CNN}. In: Lecture Notes in Computer Science (including subseries Lecture Notes in Artificial Intelligence and Lecture Notes in Bioinformatics). vol. 10008 LNCS, pp. 121--129. Springer, Cham (2016)

\bibitem{Kurisinkel2021}
Kurisinkel, L.J., Aw, A.T., Chen, N.F.: {Coherent and Concise Radiology Report Generation via Context Specific Image Representations and Orthogonal Sentence States}. In: Conference of the North American Chapter of the Association for Computational Linguistics: Human Language Technologies (2021)

\bibitem{Lu2017}
Lu, J., Xiong, C., Parikh, D., Socher, R.: {Knowing when to look: Adaptive attention via a visual sentinel for image captioning}. In: Proceedings - 30th IEEE Conference on Computer Vision and Pattern Recognition, CVPR 2017. vol. 2017-Janua, pp. 3242--3250. Institute of Electrical and Electronics Engineers Inc. (dec 2017)

\bibitem{Ma2020}
Ma, S., Huang, Y., Che, X., Gu, R.: {Faster RCNN-based detection of cervical spinal cord injury and disc degeneration}. Journal of Applied Clinical Medical Physics  \textbf{21}(9),  235--243 (2020). \doi{10.1002/acm2.13001}

\bibitem{Mikolov2013}
Mikolov, T., Chen, K., Corrado, G., Dean, J.: {Efficient estimation of word representations in vector space}. In: 1st International Conference on Learning Representations, ICLR 2013 - Workshop Track Proceedings (2013)

\bibitem{Moon2022}
Moon, J.H., Lee, H., Shin, W., Kim, Y.H., Choi, E.: {Multi-Modal Understanding and Generation for Medical Images and Text via Vision-Language Pre-Training}. IEEE Journal of Biomedical and Health Informatics  \textbf{26}(12),  6070--6080 (dec 2022)

\bibitem{Nasrullah2019}
Nasrullah, N., Sang, J., Alam, M.S., Mateen, M., Cai, B., Hu, H.: {Automated lung nodule detection and classification using deep learning combined with multiple strategies}. Sensors (Switzerland)  \textbf{19}(17) (2019)

\bibitem{Ni2020}
Ni, J., Hsu, C.N., Gentili, A., McAuley, J.: {Learning visual-semantic embeddings for reporting abnormal findings on chest x-rays}. In: Findings of the Association for Computational Linguistics Findings of ACL: EMNLP 2020. pp. 1954--1960 (2020)

\bibitem{gpt4o}
{OpenAI}: Gpt-4o contributions (2024), \url{https://openai.com/gpt-4o-contributions/}, accessed: 2024-08-07

\bibitem{Ovacll2020}
Ovacıllı, S., Atacan, S.E., G{\"{o}}kg{\"{o}}z, G., Y{\"{u}}ksel, M., Ko{\c{c}}, O., Yıldız, A.N.: {International classification of the pneumoconiosis radiograph reader training in Turkey}. Turkish Thoracic Journal  \textbf{21}(5),  314--321 (2020)

\bibitem{Papineni2002}
Papineni, K., Roukos, S., Ward, T., Zhu, W.J.: {BLEU: A method for automatic evaluation of machine translation}. In: Proceedings of the Annual Meeting of the Association for Computational Linguistics. vol. 2002-July, pp. 311--318 (2002)

\bibitem{Radford2021}
Radford, A., Kim, J.W., Hallacy, C., Ramesh, A., Goh, G., Agarwal, S., Sastry, G., Askell, A., Mishkin, P., Clark, J., Krueger, G., Sutskever, I.: {Learning Transferable Visual Models From Natural Language Supervision}. In: Proceedings of Machine Learning Research. vol.~139, pp. 8748--8763 (2021)

\bibitem{Ren2017}
Ren, S., He, K., Girshick, R., Sun, J.: {Faster R-CNN: Towards Real-Time Object Detection with Region Proposal Networks}. IEEE Transactions on Pattern Analysis and Machine Intelligence  \textbf{39}(6),  1137--1149 (2017)

\bibitem{Robinson2016}
Robinson, J.W., Brennan, P.C., Mello-Thoms, C., Lewis, S.J.: {Reporting instructions significantly impact false positive rates when reading chest radiographs}. European Radiology  \textbf{26}(10),  3654--3659 (oct 2016)

\bibitem{Sloan2024}
Sloan, P., Clatworthy, P., Simpson, E., Mirmehdi, M.: {Automated Radiology Report Generation: A Review of Recent Advances}. IEEE Reviews in Biomedical Engineering  (2024)

\bibitem{Tanida2023}
Tanida, T., M{\"{u}}ller, P., Kaissis, G., Rueckert, D.: {Interactive and Explainable Region-guided Radiology Report Generation}. In: Proceedings of the IEEE Computer Society Conference on Computer Vision and Pattern Recognition. vol. 2023-June, pp. 7433--7442. IEEE Computer Society (apr 2023)

\bibitem{Vedantam2015}
Vedantam, R., Zitnick, C.L., Parikh, D.: {CIDEr: Consensus-based image description evaluation}. In: Proceedings of the IEEE Computer Society Conference on Computer Vision and Pattern Recognition. vol. 07-12-June, pp. 4566--4575 (2015)

\bibitem{Vinyals2015}
Vinyals, O., Toshev, A., Bengio, S., Erhan, D.: {Show and tell: A neural image caption generator}. In: Proceedings of the IEEE Computer Society Conference on Computer Vision and Pattern Recognition. vol. 07-12-June, pp. 3156--3164 (2015)

\bibitem{Visser2020}
Visser, J.J., de~Vries, M., Kors, J.A.: {Assessment of actionable findings in radiology reports}. European Journal of Radiology  \textbf{129} (2020)

\bibitem{Wang2022}
Wang, S., Tang, L., Lin, M., Shih, G., Ding, Y., Peng, Y.: {Prior Knowledge Enhances Radiology Report Generation}. AMIA Symposium  (2022)

\bibitem{Wang2021}
Wang, Z., Zhou, L., Wang, L., Li, X.: {A Self-boosting Framework for Automated Radiographic Report Generation}. In: Proceedings of the IEEE Computer Society Conference on Computer Vision and Pattern Recognition. pp. 2433--2442 (2021)

\bibitem{Wu2021}
Wu, J., Agu, N., Lourentzou, I., Sharma, A., Paguio, J., Yao, J.S., Dee, E.C., Mitchell, W., Kashyap, S., Giovannini, A., Celi, L.A., Syeda-Mahmood, T., Moradi, M.: Chest imagenome dataset (version 1.0.0) (2021), version 1.0.0

\bibitem{Wu2024}
Wu, J., Guo, D., Wang, G., Yue, Q., Yu, H., Li, K., Zhang, S.: {FPL+: Filtered Pseudo Label-based Unsupervised Cross-Modality Adaptation for 3D Medical Image Segmentation}. IEEE Transactions on Medical Imaging  \textbf{XX}, ~1 (2024)

\bibitem{Wu2023b}
Wu, T.W., Huang, J.H., Lin, J., Worring, M.: {Expert-defined Keywords Improve Interpretability of Retinal Image Captioning}. In: Proceedings - 2023 IEEE Winter Conference on Applications of Computer Vision, WACV 2023. pp. 1859--1868 (2023)

\bibitem{Xiong2019}
Xiong, Y., Du, B., Yan, P.: {Reinforced Transformer for Medical Image Captioning}. In: Lecture Notes in Computer Science (including subseries Lecture Notes in Artificial Intelligence and Lecture Notes in Bioinformatics). pp. 673--680. Springer (2019)

\bibitem{Xu2021}
Xu, J.: {A Review of Self-supervised Learning Methods in the Field of Medical Image Analysis}. International Journal of Image, Graphics and Signal Processing  \textbf{13}(4),  33--46 (aug 2021). \doi{10.5815/ijigsp.2021.04.03}

\bibitem{Yan2019}
Yan, F., Huang, X., Yao, Y., Lu, M., Li, M.: {Combining LSTM and DenseNet for Automatic Annotation and Classification of Chest X-Ray Images}. IEEE Access  (2019)

\bibitem{Yan2022}
Yan, S.: {Memory-aligned Knowledge Graph for Clinically Accurate Radiology Image Report Generation}. In: Proceedings of the Annual Meeting of the Association for Computational Linguistics. pp. 116--122 (2022)

\bibitem{You2023}
You, K., Gu, J., Ham, J., Park, B., Kim, J., Hong, E.K., Baek, W., Roh, B.: {CXR-CLIP: Toward Large Scale Chest X-ray Language-Image Pre-training}. In: Lecture Notes in Computer Science (including subseries Lecture Notes in Artificial Intelligence and Lecture Notes in Bioinformatics). pp. 101--111 (2023)

\bibitem{Zhang2022}
Zhang, D., Ren, A., Liang, J., Liu, Q., Wang, H., Ma, Y.: {Improving Medical X-ray Report Generation by Using Knowledge Graph}. Applied Sciences (21) (2022)

\bibitem{Zhang2024}
Zhang, T., Wei, D., Zhu, M., Gu, S., Zheng, Y.: {Self-supervised learning for medical image data with anatomy-oriented imaging planes}. Medical Image Analysis  (2024)

\bibitem{Ziegelmayer2023}
Ziegelmayer, S., Marka, A.W., Lenhart, N., Nehls, N., Reischl, S., Harder, F., Sauter, A., Makowski, M., Graf, M., Gawlitza, J.: {Evaluation of GPT-4's Chest X-Ray Impression Generation: A Reader Study on Performance and Perception}. Journal of Medical Internet Research  \textbf{25}(1) (2023)

\end{thebibliography}

\newpage

\appendix

\section{Appendix: Parameter Settings}
\label{appendix: parameter_settings}

Table \ref{tab: report_generation_model_hyperparameter_settings}, \ref{tab: report_generation_model_training_parameters} show the hyperparameter settings and training parameter settings of the report generation module.

\begin{table}[h]
    \centering
    \scalebox{.85}{
    \begin{tabularx}{\textwidth}{XXX}
    \toprule
    \textbf{Hyperparameter} & \textbf{RRG-base} & \textbf{RRG-opt} \\ \midrule
    num\_classes & 30 & 30 \\
    pos\_weight & 1 & 2 \\
    token\_num & 500 & 300 \\
    \bottomrule
    \end{tabularx}}
    \caption{Hyperparameter Settings for Radiology Report Generation Module.}
    \label{tab: report_generation_model_hyperparameter_settings}

    \scalebox{.7}{
    \begin{tabular}{@{}l*{6}{p{1.5cm}}@{}}
    \toprule
    \textbf{Training Parameter} & \multicolumn{3}{l}{\textbf{RRG-base}} & \multicolumn{3}{l}{\textbf{RRG-opt}} \\ \cmidrule(lr){2-4} \cmidrule(lr){5-7}
    & Object Detection & Region Selection & Generation Submodule & Object Detection & Region Selection & Generation Submodule \\ \midrule
    epochs & 2 & 2 & 2 & 10 & 10 & 5 \\
    batch\_size & 16 & 16 & 8 & 16 & 16 & 1 \\
    learn\_rate & 0.01 & 0.0001 & 0.0001 & 0.001 & 0.00005 & 0.00005 \\
    factor\_learn\_rate\_scheduler & 0.3 & 0.3 & 0.3 & 0.5 & 0.5 & 0.5 \\
    cooldown\_learn\_rate\_scheduler & 5 & 5 & 5 & 5 & 5 & 5 \\
    patience\_learn\_rate\_scheduler & 5 & 10 & 10 & 5 & 5 & 3 \\
    \bottomrule
    \end{tabular}}
    \caption{Training Parameter Settings for Radiology Report Generation Module.}
    \label{tab: report_generation_model_training_parameters}
\end{table}

Table \ref{tab: radiograph_classification_hyperparameters}, \ref{tab: radiograph_classification_training_params} show the hyperparameter settings and training parameter settings of the radiograph classification module.

\begin{table}[h]
    \centering
    \scalebox{.85}{
    \begin{tabularx}{\textwidth}{@{}l*{4}{X}@{}}
    \toprule
    \textbf{Hyperparameter} & \multicolumn{2}{l}{\textbf{R-CLIP-base}} & \multicolumn{2}{l}{\textbf{R-CLIP-opt}} \\ \cmidrule(lr){2-3} \cmidrule(lr){4-5}
    & CLIP Backbone & Downstream Classifier & CLIP Backbone & Downstream Classifier \\ \midrule
    image\_to\_image\_loss\_weight & 1 & 1 & 1 & 1 \\
    text\_to\_text\_loss\_weight & 0.5 & 0.5 & 0.5 & 0.5 \\
    loss\_ratio & 1 & 1 & 1 & 1 \\
    \bottomrule
    \end{tabularx}}
    \caption{Hyperparameter Settings for Radiograph Classification Module.}
    \label{tab: radiograph_classification_hyperparameters}

    \centering
    \scalebox{.85}{
    \begin{tabularx}{\textwidth}{@{}l*{4}{X}@{}}
    \toprule
    \textbf{Training Parameter} & \multicolumn{2}{l}{\textbf{R-CLIP-base}} & \multicolumn{2}{l}{\textbf{R-CLIP-opt}} \\ \cmidrule(lr){2-3} \cmidrule(lr){4-5}
    & CLIP Backbone & Downstream Classifier & CLIP Backbone & Downstream Classifier \\ \midrule
    batch\_size & 32 & 32 & 32 & 32 \\
    learn\_rate & 0.00005 & 0.00005 & 0.00005 & 0.00005 \\
    weight\_decay & 0.0001 & 0.0001 & 0.0001 & 0.0001 \\
    total\_epochs & 5 & 5 & 5 & 5 \\
    warmup\_epochs & 1 & 1 & 1 & 1 \\
    \bottomrule
    \end{tabularx}}
    \caption{Training Parameter Settings for Radiograph Classification Module.}
    \label{tab: radiograph_classification_training_params}
\end{table}

\end{document}